\documentclass[APA,LATO1COL]{WileyNJD-v2}

\articletype{Article Type}%

\received{26 April 2016}
\revised{6 June 2016}
\accepted{6 June 2016}

\RequirePackage{algorithm}
\RequirePackage{adjustbox}
\RequirePackage{multidef}
\RequirePackage{enumerate}
\RequirePackage{bbm}
\RequirePackage{wrapfig}
\RequirePackage{float}
\RequirePackage{makecell}

\renewcommand{\bar}[1]{\mkern 1.5mu\overline{\mkern-1.5mu#1\mkern-1.5mu}\mkern 1.5mu}
\newcommand{\ind}{\mathbbm{1}}
\newcommand{\op}[1]{\mathsf{#1}}

\DeclareMathOperator*{\argmax}{argmax}

\multidef[prefix=c]{\ensuremath{\mathcal{#1}}}{A-Z}
\multidef[prefix=cc]{\ensuremath{\mathscr{#1}}}{A-Z}
\multidef[prefix=b]{\ensuremath{\boldsymbol{#1}}}{A-Z}
\multidef[prefix=bb]{\ensuremath{\mathbbm{#1}}}{A-Z}
\multidef[prefix=b]{\ensuremath{\boldsymbol{#1}}}{a,r,x}

\newcommand{\btheta}{\boldsymbol{\theta}}
\newcommand{\brho}{\boldsymbol{\varrho}}
\def\rhoconsm{\rho^{-}_{\mathsf{C}}}
\def\rhoconsp{\rho^{+}_{\mathsf{C}}}
\def\rhoambp{\rho^{+}_{\mathsf{A}}}
\def\rhoambm{\rho^{-}_{\mathsf{A}}}
\def\alphaconsm{\alpha^{-}_{\mathsf{C}}}
\def\alphaconsp{\alpha^{+}_{\mathsf{C}}}
\def\alphaambp{\alpha^{+}_{\mathsf{A}}}
\def\alphaambm{\alpha^{-}_{\mathsf{A}}}
\def\betaconsm{\beta^{-}_{\mathsf{C}}}
\def\betaconsp{\beta^{+}_{\mathsf{C}}}
\def\betaambp{\beta^{+}_{\mathsf{A}}}
\def\betaambm{\beta^{-}_{\mathsf{A}}}
\def\contp{\mathsf{CTP}}
\def\contn{\mathsf{CTN}}
\def\confp{\mathsf{CFP}}
\def\confn{\mathsf{CFN}}
\def\aap{\mathsf{AAP}}
\def\pap{\mathsf{PAP}}
\def\aan{\mathsf{AAN}}
\def\pan{\mathsf{PAN}}
\def\tp{\mathsf{TP}}
\def\tn{\mathsf{TN}}
\def\fp{\mathsf{FP}}
\def\fn{\mathsf{FN}}

\begin{document}

\title{SOAR: Simultaneous Or-of-And Rules for Classification of Positive \& Negative Classes}

\author[1]{Elena Khusainova*}

\author[1]{Emily Dodwell}

\author[1]{Ritwik Mitra}

\authormark{Elena Khusainova, Emily Dodwell, Ritwik Mitra}

\address[1]{CDO,\\ AT\&T Data Science \& AI Research,\\ NY, USA}



\corres{*AT\&T, 33 Thomas St, New York, \\
New York, 10007, USA \\\email{elena.r.khusainova@gmail.com}}


\abstract[]{
Algorithmic decision making has proliferated and now impacts our daily lives in both mundane and consequential ways. 
Machine learning practitioners use a myriad of algorithms for predictive models in applications as diverse as movie recommendations, medical diagnoses, and parole recommendations without delving into the reasons driving specific predictive decisions. 
The algorithms in such applications are often chosen for their superior performance among a pool of competing algorithms; however popular choices such as random forest and deep neural networks fail to provide an interpretable understanding of the model's predictions.  
In recent years, rule-based algorithms have provided a valuable alternative to address this issue. \citet{wang2015bayesian} and \citet{wang2017bayesian} presented an or-of-and (disjunctive normal form) based classification technique that allows for classification rule mining of a single class in a binary classification. 
In this work, we extend this idea to provide classification rules for both classes simultaneously. 
That is, we provide a distinct set of rules for each of the positive and negative classes. 
We also present a novel and complete taxonomy of classifications that clearly capture and quantify the inherent ambiguity of noisy binary classifications in the real world. We show that this approach leads to a more granular formulation of the likelihood model and a simulated annealing-based optimization achieves classification performance competitive with comparable techniques.
We apply our method to synthetic and real world data sets for comparison with other related methods to demonstrate the utility of our contribution.
}



\maketitle


\section{Introduction}\label{sec:intro}
The use of machine learning (ML) tools is ubiquitous in today's world, with applications ranging from the purely technical, e.g. facial recognition or spam detection, to the biomedical (\citet{shen2015machine}, \citet{esteva2017dermatologist}, \citet{libbrecht2015machine}) and social sciences (\citet{berk2012criminal}). As researchers and practitioners turn to ML methods to perform increasingly complicated and high-impact tasks, good model fit and high predictive accuracy of ML tools -- while desirable -- are proving insufficient. There is a need for transparency in large-scale automated decisions to enable proper assignment of accountability for applications with important implications to human lives; see \citet{varshney2017safety}. This need has found further validation in the European Union's (EU) adoption of General Data Protection Regulation (GDPR)  (see \cite{gdpr2016}), guidelines that address the collection, storage, processing, and use of personal data of EU residents. GDPR stipulates that in situations where personal data are processed for use in an automated decision mechanism, individuals should have 
\begin{quote}
{\it the right to obtain human intervention, to express his or her point of view, to obtain an explanation of the decision reached after such assessment and to challenge the decision.}
\end{quote}
This could potentially lead the way for wider adoption of and preference for interpretable, non-black-box models in certain domains. See also \citet{goodman2017european} and \citet{wachter17} for further discussion of legal implications regarding interpretability and explainability as stated in GDPR.

This need for understanding has spurred the development of Interpretable Machine Learning (IML), a sub-division of machine learning that focuses on discovering and making explicit the relationships between predictors and a specific outcome in a form understandable by humans. \textit{Interpretability} in machine learning is a rapidly growing field\footnote{Google Scholar produces more than 5000 results for the search ``Interpretable Machine Learning'' for the period from January 2015 to April 2020 as opposed to less than 200 for all time before 2015.}, and researchers have tried to achieve it in a variety of ways. Perhaps necessarily, it is an ambiguous term with `domain-specific' implications (\citet{rudin2019stop}, \citet{murdoch2019interpretable}, \citet{doshi2017towards}). One approach to achieve interpretability has been to build black box models and assign explainability to their predictions post-hoc through various mechanisms; we refer to \citet{guidotti2018survey} for a detailed accounting and references. In this current work, we instead adopt IML to refer to that definition suggested by \citet{murdoch2019interpretable}:
\begin{quote}
\textit{The use of machine-learning models for the extraction of relevant knowledge about domain relationships contained in data.} 
\end{quote}
The term {\it relevant} means that the insights obtained are useful for the chosen audience and problem domain.
The interpretability that we are interested in is {\it model-based}, rather than {\it post hoc}, meaning that we do not seek new ways of deriving insights from existing methods, but rather create a new algorithm that is inherently designed to be interpretable. 

Given this increased need for transparency in situations where ML algorithms have the potential to significantly impact human subjects, and in light of the distinction between explainable and interpretable models, we focus our attention on the latter with the evolution of rule-based classifiers that are inherently interpretable.  

A widely-used class of IML tools is rule ensemble algorithms (\citet{friedman2008predictive}).
Rule-based algorithms are inherently interpretable as they fit the model with a set of rules that can be understood by a human.
For example, the following may define the outcome of a customer's choice regarding whether or not to buy a piece of clothing based on its color, size, and price:
\begin{itemize}
    \item[] If {\it color} is {\it blue} and  {\it price} is {\it low} then {\it buy = ``yes''} 
    \item[] If {\it color} is {\it red} and  {\it size} is {\it M} and {\it price} is {\it high} then {\it buy = ``no''} 
    \item[] If {\it size} is {\it XXXS}  then {\it buy = ``yes''} 
\end{itemize}
Such rules enable us to see relationships between variables by expressing dependencies of the outcome variable on the predictors in a manner that can be easily understood. 

The history of rule-based interpretable machine learning (especially rules of the above form) is rich, and the field is growing. Early instances of rule-based ML models may be found in \citet{crama1988cause} and \citet{boros2000implementation} in the context of ``logical analysis of data,'' which aims to detect and group patterns that can correctly classify examples. \citet{friedman1999bump} introduced the Patient Rule Induction Method (PRIM), which identifies simple rules to define rectangular subregions of the input variable space that correspond to above- or below-average values of the output variable. Further extensions of PRIM have been studied in  \citet{goh2014box}, where exact and approximate-yet-fast rule sets can be derived through mixed integer programming. 	\citet{cohen1995fast} proposed a fast algorithm RIPPER$k$ which creates a rule-based model by first growing a rule set and then repeatedly pruning it based on error-reducing criteria. We refer to \citet{molnar2020interpretable} for a recent accounting of interpretable machine learning methods and further references because research in this field has evolved in various directions too numerous to note here in detail.

Relevant to our current work, \citet{wang2015bayesian} proposed a rule-based algorithm for binary classification called Bayesian Or's of And's (BOA), which represents a Bayesian framework to address the issue of model interpretability.  
As the name suggests, the algorithm requires a prior distribution on the set of possible rules that may serve as a proxy for professional opinion about the shape of the rule set. This algorithm results in one rule set that defines the positively classified observations, while the negative class is passively defined as its complement.
The algorithm was created in the context of recommendation systems (\citet{wang2015bayesian}) for which the positive class is of primary interest.

However, in certain applications, obtaining a rule set for the negative class may be just as important as defining the positive class. Consider for example, a classification model that predicts whether a patient will have a drug-related adverse reaction (positive class) or not (negative class) given their medical history, including e.g. comorbidities, results of diagnostic tests, and prior prescribed medication. Such a model may enable safer administration of a newer drug when, for example, adverse effects of a treatment and its potential toxicity require further study and consideration. In this scenario, there is significant value for both the doctor and subject receiving the treatment to understand the combination of factors associated with non-occurrence of an adverse event (i.e. definition of the negative class) to inform safe treatment and motivate patient compliance. Furthermore, cases where rules associated with both an adverse event's occurrence and non-occurrence are satisfied may indicate complicated interactions of multiple factors, which alerts both the doctor and patient to the potential for extenuating circumstances and enable closer monitoring. We address this important aspect of rule-based modeling directly in this work.  
\vskip5pt
\noindent\textbf{Contribution.} In this paper we propose SOAR, {\bf S}imultaneous {\bf O}r-of-{\bf A}nd {\bf R}ules, an algorithm to identify rule sets for each of the positive and negative classes in a binary classification problem. 
The SOAR algorithm is an extension of the BOA algorithm. It treats the positive and negative classes separately and produces a defined rule set for each, rather than creating rules for the positive class and assigning the complement to the negative class. We therefore take a direct approach through the Bayesian framework as proposed in \citet{wang2015bayesian} and describe a complete likelihood for both classes.  We calculate the posterior likelihood through annealing (\citet{van1987simulated}). A key advantage of this approach is that two rule sets allow for cases that may satisfy both or neither of them; we refer to such cases as ``ambiguous''. We argue that for these ambiguous cases, insight into which of an observation's traits impact its assignment to both the positive and negative classes (or neither) is important for interpretability.
Recognizing this need for describing ambiguity, SOAR provides a complete \textit{taxonomy} of classifications that identify conflicts when the same rules suggest different classes. This taxonomy is useful and important for real world applications which tend to be noisy and do not lend themselves to neat categorizations.\\
\indent \textcolor{black}{Given the objective of SOAR, some comparison to the literature on uncertainty quantification (UQ) is unavoidable. The objective of UQ is usually to model uncertainty in a probabilistic way, often through Bayesian modeling, taking into account two main sources of uncertainty in machine learning models: aleatoric and epistemic; see \citet{der2009aleatory}, \citet{hullermeier2021aleatoric}. In broad terms, the goal of UQ is assign a measure of plausibility to an outcome based on the model uncertainty (epistemic) and data noise (aleatoric). The objective of SOAR is not to assign a specific metric of uncertainty, but rather to clearly delineate the ambiguity in the outcomes through interpretable rule sets. It is conceivable that some measure of uncertainty can be assigned to the prediction outcomes from SOAR based on the posterior probabilities of the rule sets. We do not pursue this line of research in our current work, but rather leave it for future exploration.}\\
\indent Our work is organized as follows: in Section \ref{sec:problem}, we present an implementation of the SOAR algorithm and discuss terminology and key notation as they appear throughout the remaining sections. We present a detailed description of our proposal and introduce the complete taxonomy of classifications in Section \ref{sec:algorithm}, as well as explain the algorithm for creating simultaneous rules. In Section \ref{sec:results}, we validate our algorithm with one simulated and two real-word data sets. By comparing SOAR to algorithms such as BOA and random forest, we find that SOAR's performance is comparable in predictive accuracy; by offering deeper insights provided by a rule set that explicitly defines the negative class, our algorithm provides increased descriptive accuracy.

\section{Problem set up}\label{sec:problem}
In this section we define the problem and the terminology we use.
When possible we adopt notation and terminology from either \citet{friedman2008predictive} or \citet{wang2015bayesian}.\\
\indent The algorithm is intended to address the standard binary classification problem. 
We assume that the data, denoted as $\bX$, consist of $p$ categorical predictors and $n$ observations.
That is  $\bX:=[\bX_{\boldsymbol{\cdot}1},...,\bX_{\boldsymbol{\cdot}p}]$, with ${\bX_{\boldsymbol{\cdot}i}\in\{0,1\}^n}$ for $1\leq i\leq p$.
Note, that $\bX$ can be also expressed as $\bX:=[\bX_{1\boldsymbol{\cdot}},...,\bX_{n\boldsymbol{\cdot}}]^T$.
When it does not cause confusion we will abandon dots in the notation and use $\bx_i=\bX_{i\boldsymbol{\cdot}}$ to refer to observations in the data.
For simplicity of notation, we also say that $\bX=\{\bx_1,...,\bx_n\}$.
Additionally, each observation belongs to one of two classes, labeled as $0$ and $1$, with labels stored in the variable of interest $Y=[y_1,...,y_n]\in\{0,1\}^n$.\\
\indent In order to define the form of the model to fit the data, we need to first define rules. 
Let function $\br=\br(\bx)$ be a product of indicators:
\begin{align}\label{eq:ruleDef}
    \br(\bx) = \prod\limits^{k}_{i=1}\ind\{\bx[s_i]=x_{i}\},
\end{align}
where $\bx\in\bX$, $1\leq s_i\leq p$ and values $x_i$ belong to the range of corresponding predictors $\bX_{\boldsymbol{\cdot} s_i}$, $i=1,...,k$.
We say that any function of the form (\ref{eq:ruleDef}) is a {\it rule} or {\it pattern} of length $k$.
For a set of rules $\bR$ we say that 
\begin{align}
    \bR(\bx)
    =\begin{cases}
    1 & \text{ if there exists }  \br\in \bR \text{ such that } \br(\bx)=1,\\
    0 & \text{otherwise}
\end{cases}
\end{align}
and define 
$
    \bX_{\bR}:=\{\bx\in\bX:~\bR(\bx)=1\}
$
to be the set of all observations that satisfy the rule set $\bR$.\\
\indent The model we fit then takes form:
\begin{align*}
    f(\bx)=\sum\limits^{M}_{m=1}\br_m(\bx),
\end{align*}
where $\br_1,...,\br_m$ are rules.
This form of the model can be seen as a disjunction of conjunctions, or simply, Or-of-Ands (hence the name of the algorithm).
If sets $\bX^+$ and $\bX^-$ are such that 
\begin{align*}
    \bX^+=\,\{\bx_i\in\bX:~y_i=f(\bx_i)=1\},\quad
    \bX^-=\,\{\bx_i\in\bX:~y_i=f(\bx_i)=0\},
\end{align*}
that is, $\bX^+$ and $\bX^-$ consist of observations from positive and negative classes respectively, then the goal of the algorithm is to find rule sets $\bR^+$ and $\bR^-$ such that $\bX_{\bR^+}$ and $\bX_{\bR^-}$ are close (in terms of maximizing the posterior distribution, see Section \ref{sec:priorAndModel}) to $\bX^+$ and $\bX^-$ respectively.

\section{Simultaneous Or-of-And Rules}\label{sec:algorithm}

In this section we present the construction of the prior and log-likelihood model and also introduce the new taxonomy to be used with simultaneous classification.\\
\indent The main contribution of the SOAR algorithm is that instead of finding trends and patterns among just the positive observations, we do so simultaneously for each of the positive and negative classes. 

\subsection{Taxonomy of classifications}

Consider two rule sets $\bR^+_*$ and $\bR^-_*$ and the corresponding sets of observations $\bX_{\bR^+_*}$ and $\bX_{\bR^-_*}$. Intuitively, $\bR^+_*$ and $\bR^-_*$ describe the characteristics of positive and negative sets respectively. 
We call cases in $\bX_{\bR^+_*}\cap\bX_{\bR^-_*}$ {\it active ambiguous} and those in $\bX\setminus\left(\bX_{\bR^+_*}\cup\bX_{\bR^-_*}\right)$ {\it passive ambiguous}. That is, active ambiguous refers to those cases that exhibit characteristics of both classes, while passive ambiguous are cases that do not exhibit characteristics of either class.\\
\indent For cases in $\bX_{\bR^+_*}\cap\bar{\bX_{\bR^-_*}}$ or $\bX_{\bR^-_*}\cap\bar{\bX_{\bR^+_*}}$, where $\bar{\bX_{\bR^-_*}}:=\bX\setminus\bX_{\bR^-_*}$ and ${\bar{\bX_{\bR^+_*}}:=\bX\setminus\bX_{\bR^+_*}}$, we say that there is {\it consensus}.
There represent cases that satisfy the positive class definition and not that of the negative class, or vice versa.
The complete taxonomy is presented in Table \ref{tb:taxonomy}.
For an observation $\bx_i$ we can have one and only one of the eight possible scenarios.
The first four cases are standard for classification problems, but the last four are slightly unusual. We allow the algorithm to not give any prediction if there is uncertainty in the data which in turn highlights the inherent ambiguity.
We argue that allowing such cases makes our method more appropriate in circumstances where the user would prefer a second opinion model, as our expanded taxonomy provides for this detailed look.

It is worthwhile to point out the differences between our notations and more well known measures such as positive predictive (PPR) and true positive rate (TPR). Recall that PPR = TP/(all discovered positives) and TPR = TP/(all positives in the data). Our definition states that it is true positive among all the positive decisions from “both rule sets”. Because we are focusing on ambiguous cases, the binarized language of TPR and PPR doesn’t transfer directly to our setup.


\subsection{Beta-Binomial prior and log-likelihood model}\label{sec:priorAndModel}

We now describe the prior distribution for the model and the model itself.
The first step of the algorithm (described in Section \ref{sec:ATTalgo}) results in two pattern sets $\cP^+$ and $\cP^-$ called {\it pattern pools}.
The patterns in each of them satisfy three conditions: 
\begin{itemize}
    \item each pattern consists of no more than $L$ literals, with $L$ provided by the user;
    \item each pattern is frequent in the data. That is,  there are more than a pre-specified threshold of observations that agree with the pattern;
    \item each pattern on its own has a strong relationship with the outcome variable (in terms of impurity score).
\end{itemize}
The prior is defined on the sets of all subsets of $\cP^+$ and $\cP^-$ in the following way:
\begin{enumerate}
    \item For each $l^+,l^-\in\{1,2,...,L\}$ we simulate $p_l^+$ and $p_l^-$ such that
    \[
        p_{l^+}^+\sim \op{Beta}(\alpha^+_l,\beta^+_l)
        ~~\mbox{ and }~~
        p_{l^-}^-\sim \op{Beta}(\alpha^-_l,\beta^-_l).
    \]
    The parameters 
    $
    \btheta_{prior}=\{\alpha^+_1,...,\alpha^+_{L^+},\alpha^-_1,...,\alpha^-_{L^-}, \beta^+_1,...,\beta^+_{L^+}, \beta^-_1,...,\beta^-_{L^-}\}
    $
    are user-specified.
    \item Each pattern of length $l^+$ (or $l^-$) is selected from $\cP^+$ (or $\cP^-$) with probability $p^+_{l^+}$ (or $p^-_{l^-}$).
\end{enumerate}
In subsequent discussions, we describe the prior on the positive and negative rules jointly as 
$
    \pi\Big(\bR^+, \bR^- \ | \  \btheta_{prior}\Big).
$

\indent With full control over parameters of the prior, the user can influence the shape of the outcome by limiting the maximal pattern length, favoring longer or shorter patterns, and affecting the size of each of the rule sets.\\
\indent We next explicitly write the likelihood of the data given a model.
The following likelihood parameters govern the probability that an observation is of a true positive (or negative) class when it satisfies the corresponding pattern sets:
\begin{align}\label{eq:rho_parameters}
 \rhoambp  =  \bbP(y_{n} = 1\ |\ \bR^+(\bx_{n}) = 1, \bR^-(\bx_{n}) = 1 ),
 \quad\rhoambm  =  \bbP(y_{n} = 0\ |\ \bR^+(\bx_{n}) = 0, \bR^-(\bx_{n}) = 0 ),\notag\\
 \rhoconsp  =  \bbP(y_{n} = 1\ |\ \bR^+(\bx_{n}) = 1, \bR^-(\bx_{n}) = 0 ),
 \quad\rhoconsm  =  \bbP(y_{n} = 0\ |\ \bR^+(\bx_{n}) = 0, \bR^-(\bx_{n}) = 1 ).
\end{align}
The prior for these four parameters are assigned as follows:
\begin{align*}
 \rhoambp \sim \op{Beta}(\alphaambp, \betaambp),\quad
 \rhoambm \sim \op{Beta}(\alphaambm, \betaambm),\quad
 \rhoconsp \sim \op{Beta}(\alphaconsp, \betaconsp),\quad
 \rhoconsm \sim \op{Beta}(\alphaconsm, \betaconsm).
\end{align*}
As before, parameters $\btheta_{likelihood}=\{\alphaambp, \betaambp, \alphaambm, \betaambm, \alphaconsp, \betaconsp, \alphaconsm, \betaconsm\}$ are provided by the user.
Denote $\brho = \{\rhoambp, \rhoambm, \rhoconsp, \rhoconsm\}$; then the log-likelihood of the data is given by:
\begin{align}\label{eq:lik1}
 & \log\big(\bbP(\{ \bx_i, y_i\}^{n}_{i=1}\ | \ \bR^+, \bR^-, \brho)\big) \nonumber\\
 & =\sum_{i=1}^{n} \Bigg [\Big\{y_{i}\bR^+(\bx_i)(1-\bR^-(\bx_i))\Big\}\log\rhoconsp  + \Big\{(1-y_{i})\bR^+(\bx_i)(1-\bR^-(\bx_i))\Big\}\log(1-\rhoconsp) \notag\\
 & \qquad + \Big\{(1-y_{i})(1-\bR^+(\bx_i))\bR^-(\bx_i)\Big\}\log\rhoconsm  + \Big\{y_{i}(1-\bR^+(\bx_i))\bR^-(\bx_i)\Big\}\log(1-\rhoconsm) \notag\\
 & \qquad + \Big\{y_{i}\bR^+(\bx_i)\bR^-(\bx_i)\Big\}\log\rhoambp + \Big\{(1-y_{i})\bR^+(\bx_i)\bR^-(\bx_i)\Big\}\log(1-\rhoambp) \notag\\
 & \qquad + \Big\{(1-y_{i})(1-\bR^+(\bx_i))(1-\bR^-(\bx_i))\Big\}\log\rhoambm  + \Big\{y_{i}(1-\bR^+(\bx_i))(1-\bR^-(\bx_i))\Big\}\log(1-\rhoambm)\Bigg]\notag\\
 & = \contp \log(\rhoconsp) + \confp \log(1-\rhoconsp) + \contn \log(\rhoconsm) + \confn \log(1-\rhoconsm) \notag\\
 & \qquad + \aap \log(\rhoambp) + \aan \log(1-\rhoambp) + \pan \log(\rhoambm) + \pap\log(1-\rhoambm),
\end{align}
where $\contp, \confp, \contn, \confn, \aap, \aan, \pan, \pap$ are the counts of the corresponding cases.
Here we have (with slight abuse of notation) used the empirical versions of the definitions defined in Table~ \ref{tb:taxonomy}. Incorporating the prior structure for $\brho$, we get 
\begin{align}\label{eq:lik2}
 & \bbP(\{\bx_i, y_i\}^{n}_{i=1}\ | \bR^+, \bR^-, \btheta_{likelihood}) \nonumber\\
 & = \int_{\brho} \big(\bbP(\{ \bx_i, y_i\}^{n}_{i=1}\ |\ \bR^+, \bR^-, \brho)\big)  \times\op{Beta}(\alphaambp, \betaambp)\op{Beta}(\alphaambm, \betaambm)\op{Beta}(\alphaconsm, \betaconsm)\op{Beta}(\alphaconsp, \betaconsp)  d\brho \notag\\
 & =  \int_{\brho}(\rhoconsp)^{\contp} (1-\rhoconsp)^{\confp} (\rhoconsm)^{\contn} (1-\rhoconsm)^{\confn} \times (\rhoambp)^{\aap} (1-\rhoambp)^{\aan}(\rhoambm)^{\pan }(1-\rhoambm)^{\pap} \notag\\
 & = \Bigg\{\frac{\op{B}(\contp + \alphaconsp, \confp + \betaconsp)}{\op{B}(\alphaconsp, \betaconsp)}\Bigg\}
 \times \Bigg\{\frac{\op{B}(\contn + \alphaconsm, \confn + \betaconsm)}{\op{B}(\alphaconsm, \betaconsm)}\Bigg\}\notag\\
 & \qquad \times \Bigg\{\frac{B(\aap + \alphaambp, \aan + \betaambp)}{\op{B}(\alphaambp, \betaambp)}\Bigg\}
 \times\Bigg\{\dfrac{\op{B}(\pan + \betaambm, \pap + \alphaambm)}{\op{B}(\alphaambm, \betaambm)}\Bigg\},
\end{align}
where $\op{B}(a,b) = \Gamma(a)\Gamma(b)/\Gamma(a+b)$. The final posterior distribution for the rules and data is then given by:
\begin{align}\label{eq:posterior}
 & \bbP\Big(\{\bx_i, y_i\}^{n}_{i=1}, \bR^+, \bR^- \ | \ \btheta_{likelihood}, \btheta_{prior}\Big) \propto \bbP\Big(\{ \bx_i, y_i\}^{n}_{i=1}\ | \ \bR^+, \bR^-, \btheta_{likelihood} \Big) \times \pi\Big(\bR^+, \bR^- \ | \ \btheta_{prior}\Big).
\end{align}
The MAP (maximum {\it a posteriori}) estimator for these rules is then given by:
\begin{align}\label{eq:map}
 \big(\bR^+_*, \bR^-_*\big) 
 := \argmax_{\bR^+ \subseteq \cP^+, \bR^-\subseteq \cP^-}  \bbP\Big(\{ \bx_i, y_i\}^{n}_{i=1}, \bR^+, \bR^- \ |\ 
 \btheta_{likelihood}, \btheta_{prior}\Big).
\end{align}

\subsection{Algorithm}\label{sec:ATTalgo}

The algorithm can be divided into three main steps:
\begin{enumerate}
    \item Frequent pattern mining. The most frequent patterns are mined from  the data using the scalable FPGrowth algorithm introduced by \citet{han2000mining} and implemented by \citet{borgelt2005implementation}.\\
    {\bf Remark.} Note that frequent pattern mining (FPM) is agnostic of the positive or negative class membership of the observation. As such, one could run a single FPM on the whole dataset to get a set of frequent rules for both positive and negative classes. 
    That is, at this stage the positive and negative pattern pools are identical. 
    \item Pattern screening. The patterns mined in the previous step are ranked based on one of the following, user-chosen, impurity functions: conditional entropy or Gini index. 
    The best ones form the final pattern pools
    of the size limited by the user.
    
    \item Simulated annealing to find MAP (\ref{eq:map}). 
    For details, see Algorithm \ref{tb:algorithm}.
\end{enumerate}

Define 
\begin{align*}
    \op{Score}\big(\bR^+, \bR^-\big)
    =-\log \bbP\Big(\{\bx_i, y_i\}^{n}_{i=1}, \bR^+, \bR^- \ | \ \btheta_{likelihood}, \btheta_{prior}\Big),
\end{align*}
and the cooling schedule:
$
    T(t):=T_0/\log(1+t),
$
where the initial temperature $T_0$ is user-defined.
Then the simulated annealing algorithm to find MAP solution is as shown in Algorithm~\ref{tb:algorithm}.


The procedures $COVERMORE$ and $COVERLESS$ used in the algorithm are the same as in \citet{wang2015bayesian}:
\begin{itemize}
 \item COVERMORE\,($\bR$, $p$)
 With probability $p$, add a random pattern to $\bR$ from the corresponding pool.
 Else, evaluate the objective $\operatorname{Score}()$ for all neighboring solutions where a pattern is added to $\bR$ and choose the one with the best score.
 \item COVERLESS\,($\bR$, $p$)
 With probability $p$, remove a random pattern from $\bR$.
 Else, evaluate the objective $\operatorname{Score}()$ for all neighboring solutions where a pattern is removed from $\bR$ and choose the one with the best score.
\end{itemize}

\section{Results} \label{sec:results}

In this section we present the comparison of SOAR's performance to that of BOA by \citet{wang2015bayesian} and random forest \citep{breiman2001random}. We use three data sets to illustrate the benefits of our algorithms, as well as its difficulties: 
\begin{enumerate}
    \item Synthetic data manually generated to enable complete control over actual positive and negative rule sets.
    \item The Adult data set (\citet{kohavi1996scaling}) extracted from the 1994 Census Bureau database, which contains socioeconomic information on $\sim$32K adults.
    \item Car data from (\citet{bohanec1988knowledge}) that represents evaluation of 1728 cars based on their characteristics.
\end{enumerate}

For comparison of algorithm performance, we split the data randomly into training and test sets, train each model on the same training set, and compare model predictions on the same test set. Because SOAR allows for ambiguous classification ($\aan$, $~\aap$, $~\pan$, $~\pap$), we differentiate such {\it ambiguous} cases from those that are {\it truly misclassified}, i.e. predicted incorrectly, by our algorithm ($\confp, \confn$). 
For classical performance comparison we also include a forced prediction. That is, in the event that SOAR predicts an observation to be actively ambiguous, it is assigned to the class that has the longest rule it agrees with (unless the lengths are the same, in which case it stays ambiguous).

\subsection{Synthetic data}

We simulate a data set of 1000 observations with 5 binary predictors ($x1,x2,x3,x4,x5$) and a single binary outcome. Because there are $2^5 = 32$ possible combinations of predictors under this scheme, this naturally results in duplicate rows in the data matrix, as we may expect in real world applications when different observations share the same characteristics. The following rules determine the true classification of each observation:
\begin{align*}
    \bR^+_{truth}
    =&\{(x1, 0)\wedge (x2, 1),\quad
    (x1, 1) \wedge (x2, 1) \wedge (x3, 0), 
    \quad (x2, 0) \wedge (x3, 0) \wedge (x5, 1)\}\\
    \bR^-_{truth}
    =&\{(x1,1) \wedge (x3, 1), \quad
    (x1,0) \wedge (x2,0) \wedge (x4,0),\quad
    \, (x1,1) \wedge (x2,0) \wedge (x3,0)\},
\end{align*}
Among the 1000 observations, 500 are classified as positive and 490 are classified as negative. The intersection between these two classes (that is, observations that satisfy both positive and negative rules) is 59, and the remaining 69 observations are neither positive nor negative. We assign those that satisfy both positive and negative rules to a final positive class with probability 0.05, and those that satisfy neither to a final positive class with probability 0.5.
Table \ref{tb:synth} summarizes the data. 


We randomly split the data into a training set of 800 observations and a test set of the remaining 200 observations. Regarding interpretability, the patterns produced by SOAR algorithm are:
\begin{align*}
    \bR^+
    =\{&(x2,1), 
    (x3,0)\wedge(x5,1)\}\\
    \bR^-
    =\{&(x1,1)\wedge(x3,1),
    (x2,0)\wedge(x5,0),
    (x1,1)\wedge(x2,0)\wedge(x4,1)\wedge(x5,1),
    (x1,1)\wedge(x2,0)\wedge(x4,0)\wedge(x5,1)\}
\end{align*}

Table \ref{tb:synth.fin} illustrates that the positive and negative classes defined by these rules are close to those defined by $\bR^+_{truth}$ and $\bR^-_{truth}$; predictor combinations for which the two pairs of rule sets disagree are marked with (*).
Because both positive and negative rule sets provide a determination as to whether or not they were satisfied, we capture the determination of each rule set for observation $\bx$ as a pair $(r^+, r^-)$, where $r^+$ is the outcome of the positive rule set, and $r^-$ is the outcome of the negative rule set.
For example, an observation $\bx=[0,0,1,0,1]$ with true class $(0,1)$ and predicted class $(0,0)$ can be interpreted as 
\begin{align*}
    \bR^+_{truth}(\bx)=0, \, \bR^-_{truth}(\bx)=1  \quad
    \bR^+(\bx)=0, \, \bR^-(\bx)=0.
\end{align*}

The comparison of all three algorithms is presented in Table~\ref{tb:synth.res}.
While the overall misclassification rate is high for unforced SOAR, no observations are truly misclassified; that is, observations for which the algorithm's prediction fails to match the true label ultimately represent ``ambiguous'' cases.
Once we force our algorithm to make a prediction (SOAR-forced), the true misclassification rate stays remarkably low and the overall misclassification rate is competitive with random forest and BOA.

\subsection{Adult data}

The Adult data set (\citet{kohavi1996scaling}) is a Census Bureau data on the income of $\sim$32K adults. 
The data has 14 categorical socioeconomic predictors and 1 binary outcome, indicating whether a person makes more than \$50K a year. 
The rules for classification are not known. The positive support (that is, people who make more than \$50K) is $\sim$8K observations, and the negative support is $\sim$24K observations.
We split the data into a training set with 25K observations and test set with $\sim$7.5K observations. 
The rules produced by SOAR are as follows:
\begin{align}\label{eq:ATTAdultRules}
    \bR^+
    =\{&(\text{marital\_status}, \text{Married-civ-spouse}),
    (\text{education\_num}, 15) ,\nonumber\\
  & (\text{education},\text{Masters})\wedge(\text{sex},\text{Male})\wedge(\text{capital\_loss},0)\wedge(\text{native\_country},\text{United-States}),\nonumber\\
  & (\text{occupation},\text{Exec-managerial})\wedge(\text{education},\text{Bachelors}) \wedge(\text{sex},\text{Male})\wedge(\text{capital\_loss},0)\}\nonumber\\
    \bR^-
    =\{&(\text{education\_num}, 9),(\text{relationship}, \text{Own-child}),(\text{occupation}, \text{Other-service}),\nonumber\\
        &(\text{capital\_gain}, 0)\wedge(\text{capital\_loss},0),(\text{occupation}, \text{Adm-clerical})\wedge(\text{sex},\text{Female})\wedge(\text{capital\_gain}, 0)\}
\end{align}

SOAR predicts that people who are married, have achieved higher levels of education, and men who work in managerial positions make more money, while people with fewer years of education, and those who work in administrative or service jobs make less. These rules are intuitive based on what we know of income factors. The comparison between the three algorithms is presented in Table~\ref{tb:adult.comp}.\\
\indent As before, SOAR's misclassification rate is very low, and the performance of SOAR-forced is superior to BOA. 
To illustrate the strength of the algorithm we show examples of ambiguous---both passive and active---cases in Table \ref{tb:adult.ex}.\\
\indent \textcolor{black}{Upon the inspection of these examples, we can see what might cause the ambiguity in prediction: the first example is a white male with a graduate degree but the occupation is not clear, thus the algorithm cannot provide a consensus prediction. We argue that this case would not be possible to classify even for a human, thus the ambiguity in prediction is an accurate reflection of reality. Moreover the produced rule sets indicate exactly where the ambiguity comes from thus providing valuable insights. The other examples follow similar patterns. E.g., in the forth example we have a white unmarried female with a BA degree with a highly paid occupation. This doesn't fall into any rule set as a highly paid occupation only appears together with male gender (possibly due to the lower frequency of $(\text{occupation},\text{Exec-managerial})\wedge (\text{sex},\text{Female})$ pattern in the data).}\\
\indent Note also, that from these examples and the rules (\ref{eq:ATTAdultRules}), we immediately see that rule-based algorithms in general and SOAR in particular can be used to assess bias inherent in this data set. 
The characteristic $(\text{sex}, \text{Male})$ is present in two patterns for positive classification, and the first example of passive ambiguous classification (a woman with a managerial position and college education) is not labeled as positive.\\
\indent To illustrate our taxonomy Table \ref{tb:adult.tax} presents the comparative results between BOA and SOAR.
We see that both algorithms are better at detecting observations in the positive class. There are only 149 $\fn$-cases for BOA and 158 $\confn$-cases for SOAR. 
Observations in the negative class prove more problematic for BOA, which has more than 2000 $\fp$-cases; comparatively, SOAR classifies the majority of those same cases as ambiguous.\\
\indent Table \ref{tb:adult.freq} presents the most common patterns that appear in final positive and negative rule sets in 100 runs of the algorithm.
These common patterns are consistent with those observed in the rules (\ref{eq:ATTAdultRules}).

\subsection{Car evaluation data}

Car data from (\citet{bohanec1988knowledge}) represents evaluation of 1728 cars according to 6 categorical variables:
\begin{itemize}
    \item {\bf buyingPrice}: buying price of a car
    \item {\bf maintainPrice}: maintenance price
    \item {\bf doors}: number of doors
    \item {\bf persons}: number of passengers
    \item {\bf lugBoot}: size of a trunk
    \item {\bf safety}: safety level
\end{itemize}
The outcome variable indicates whether the car was assessed as \textit{not acceptable}, \textit{acceptable}, \textit{good} or \textit{very good}. For the purpose of binary classification, we encode the \textit{unacceptable} category as $0$ and the three remaining categories as $1$. 
The positive support is 518 observations and negative support is 1210 observations. 
We randomly split the data into a training set of 1200 observations and a test set of 528 observations. SOAR produces the following rules:
\begin{align*}
    \bR^+
    =\{&(\text{buyingPrice},\text{low})\wedge(\text{persons},4)\wedge(\text{safety},\text{high}),(\text{buyingPrice},\text{med})\wedge(\text{persons},4)\wedge(\text{safety},\text{high}),\\
    &(\text{maintainPrice},\text{low})\wedge(\text{persons},4)\wedge(\text{safety},\text{high}),(\text{maintainPrice},\text{med})\wedge(\text{persons},4)\wedge(\text{safety},\text{high}),\\
    &(\text{maintainPrice},\text{low})\wedge(\text{persons},\text{more})\wedge(\text{safety},\text{high})\}\\
    \bR^-
    =\{&(\text{persons},2),(\text{safety},\text{low}),(\text{buyingPrice},\text{high})\wedge(\text{maintainPrice},\text{vhigh}), (\text{doors},2)\wedge(\text{lugBoot},\text{small})\\
    &(\text{buyingPrice},\text{vhigh})\wedge(\text{maintainPrice},\text{vhigh}),(\text{buyingPrice},\text{vhigh})\wedge(\text{maintainPrice},\text{high})\}
\end{align*}
\indent Based on these rules, cars that are safe, spacious, and inexpensive (either in terms of buying or maintaining) are classified as {\it acceptable}, while small or expensive cars are classified as {\it unacceptable}.
The comparison of all three algorithms is presented in Table \ref{tb:car.comp}.\\
\indent Again, we see that observations truly misclassified by SOAR and SOAR-forced are few, while ambiguous cases account for most instances of overall misclassification.
We present examples in Table \ref{tb:car.ex} to illustrate how our algorithm manages to provide useful insights even when it is uncertain. \\
\indent We see that active ambiguous cases are those cars that appear small (2 doors, small trunk), but are able to fit 4 or more people. Meanwhile, passive ambiguous cases are cars with medium safety (note that all rules for positive classification include literal $(\text{safety}, \text{high})$), but they otherwise have low buying and maintenance prices and enough space, all of which we previously observed as associated with positive classification.

\bibliography{references}

\begin{thebibliography}{}

\bibitem [\protect \citeauthoryear {%
Berk%
}{%
Berk%
}{%
{\protect \APACyear {2012}}%
}]{%
berk2012criminal}
\APACinsertmetastar {%
berk2012criminal}%
\begin{APACrefauthors}%
Berk, R.%
\end{APACrefauthors}%
\unskip\
\newblock
\APACrefYear{2012}.
\newblock
\APACrefbtitle {Criminal justice forecasts of risk: A machine learning
  approach} {Criminal justice forecasts of risk: A machine learning approach}.
\newblock
\APACaddressPublisher{}{Springer Science \& Business Media}.
\PrintBackRefs{\CurrentBib}

\bibitem [\protect \citeauthoryear {%
Bohanec%
\ \BBA {} Rajkovic%
}{%
Bohanec%
\ \BBA {} Rajkovic%
}{%
{\protect \APACyear {1988}}%
}]{%
bohanec1988knowledge}
\APACinsertmetastar {%
bohanec1988knowledge}%
\begin{APACrefauthors}%
Bohanec, M.%
\BCBT {}\ \BBA {} Rajkovic, V.%
\end{APACrefauthors}%
\unskip\
\newblock
\APACrefYearMonthDay{1988}{}{}.
\newblock
{\BBOQ}\APACrefatitle {Knowledge acquisition and explanation for
  multi-attribute decision making} {Knowledge acquisition and explanation for
  multi-attribute decision making}.{\BBCQ}
\newblock
\BIn{} \APACrefbtitle {8th Intl Workshop on Expert Systems and their
  Applications} {8th intl workshop on expert systems and their applications}\
  (\BPGS\ 59--78).
\PrintBackRefs{\CurrentBib}

\bibitem [\protect \citeauthoryear {%
Borgelt%
}{%
Borgelt%
}{%
{\protect \APACyear {2005}}%
}]{%
borgelt2005implementation}
\APACinsertmetastar {%
borgelt2005implementation}%
\begin{APACrefauthors}%
Borgelt, C.%
\end{APACrefauthors}%
\unskip\
\newblock
\APACrefYearMonthDay{2005}{}{}.
\newblock
{\BBOQ}\APACrefatitle {An Implementation of the FP-growth Algorithm} {An
  implementation of the fp-growth algorithm}.{\BBCQ}
\newblock
\BIn{} \APACrefbtitle {Proceedings of the 1st international workshop on open
  source data mining: frequent pattern mining implementations} {Proceedings of
  the 1st international workshop on open source data mining: frequent pattern
  mining implementations}\ (\BPGS\ 1--5).
\PrintBackRefs{\CurrentBib}

\bibitem [\protect \citeauthoryear {%
Boros%
\ \protect \BOthers {.}}{%
Boros%
\ \protect \BOthers {.}}{%
{\protect \APACyear {2000}}%
}]{%
boros2000implementation}
\APACinsertmetastar {%
boros2000implementation}%
\begin{APACrefauthors}%
Boros, E.%
, Hammer, P\BPBI L.%
, Ibaraki, T.%
, Kogan, A.%
, Mayoraz, E.%
\BCBL {}\ \BBA {} Muchnik, I.%
\end{APACrefauthors}%
\unskip\
\newblock
\APACrefYearMonthDay{2000}{}{}.
\newblock
{\BBOQ}\APACrefatitle {An implementation of logical analysis of data} {An
  implementation of logical analysis of data}.{\BBCQ}
\newblock
\APACjournalVolNumPages{IEEE Transactions on knowledge and Data
  Engineering}{12}{2}{292--306}.
\PrintBackRefs{\CurrentBib}

\bibitem [\protect \citeauthoryear {%
Breiman%
}{%
Breiman%
}{%
{\protect \APACyear {2001}}%
}]{%
breiman2001random}
\APACinsertmetastar {%
breiman2001random}%
\begin{APACrefauthors}%
Breiman, L.%
\end{APACrefauthors}%
\unskip\
\newblock
\APACrefYearMonthDay{2001}{}{}.
\newblock
{\BBOQ}\APACrefatitle {Random forests} {Random forests}.{\BBCQ}
\newblock
\APACjournalVolNumPages{Machine learning}{45}{1}{5--32}.
\PrintBackRefs{\CurrentBib}

\bibitem [\protect \citeauthoryear {%
Cohen%
}{%
Cohen%
}{%
{\protect \APACyear {1995}}%
}]{%
cohen1995fast}
\APACinsertmetastar {%
cohen1995fast}%
\begin{APACrefauthors}%
Cohen, W\BPBI W.%
\end{APACrefauthors}%
\unskip\
\newblock
\APACrefYearMonthDay{1995}{}{}.
\newblock
{\BBOQ}\APACrefatitle {Fast effective rule induction} {Fast effective rule
  induction}.{\BBCQ}
\newblock
\BIn{} \APACrefbtitle {Machine learning proceedings 1995} {Machine learning
  proceedings 1995}\ (\BPGS\ 115--123).
\newblock
\APACaddressPublisher{}{Elsevier}.
\PrintBackRefs{\CurrentBib}

\bibitem [\protect \citeauthoryear {%
{Council of European Union}%
}{%
{Council of European Union}%
}{%
{\protect \APACyear {2016}}%
}]{%
gdpr2016}
\APACinsertmetastar {%
gdpr2016}%
\begin{APACrefauthors}%
{Council of European Union}.%
\end{APACrefauthors}%
\unskip\
\newblock
\APACrefYearMonthDay{2016}{}{}.
\newblock
\APACrefbtitle {Reg (EU) 2016/679 of the European Parliament and of the Council
  of 27 April 2016 on the protection of natural persons with regard to the
  processing of personal data and on the free movement of such data, and
  repealing Dir 95/46/EC (General Data Protection Regulation) 2016.} {Reg (eu)
  2016/679 of the european parliament and of the council of 27 april 2016 on
  the protection of natural persons with regard to the processing of personal
  data and on the free movement of such data, and repealing dir 95/46/ec
  (general data protection regulation) 2016.}
\newblock
\APACrefnote{https://eur-lex.europa.eu/eli/reg/2016/679/oj/eng}
\PrintBackRefs{\CurrentBib}

\bibitem [\protect \citeauthoryear {%
Crama%
, Hammer%
\BCBL {}\ \BBA {} Ibaraki%
}{%
Crama%
\ \protect \BOthers {.}}{%
{\protect \APACyear {1988}}%
}]{%
crama1988cause}
\APACinsertmetastar {%
crama1988cause}%
\begin{APACrefauthors}%
Crama, Y.%
, Hammer, P\BPBI L.%
\BCBL {}\ \BBA {} Ibaraki, T.%
\end{APACrefauthors}%
\unskip\
\newblock
\APACrefYearMonthDay{1988}{}{}.
\newblock
{\BBOQ}\APACrefatitle {Cause-effect relationships and partially defined Boolean
  functions} {Cause-effect relationships and partially defined boolean
  functions}.{\BBCQ}
\newblock
\APACjournalVolNumPages{Annals of Operations Research}{16}{1}{299--325}.
\PrintBackRefs{\CurrentBib}

\bibitem [\protect \citeauthoryear {%
Der~Kiureghian%
\ \BBA {} Ditlevsen%
}{%
Der~Kiureghian%
\ \BBA {} Ditlevsen%
}{%
{\protect \APACyear {2009}}%
}]{%
der2009aleatory}
\APACinsertmetastar {%
der2009aleatory}%
\begin{APACrefauthors}%
Der~Kiureghian, A.%
\BCBT {}\ \BBA {} Ditlevsen, O.%
\end{APACrefauthors}%
\unskip\
\newblock
\APACrefYearMonthDay{2009}{}{}.
\newblock
{\BBOQ}\APACrefatitle {Aleatory or epistemic? Does it matter?} {Aleatory or
  epistemic? does it matter?}{\BBCQ}
\newblock
\APACjournalVolNumPages{Structural safety}{31}{2}{105--112}.
\PrintBackRefs{\CurrentBib}

\bibitem [\protect \citeauthoryear {%
Doshi-Velez%
\ \BBA {} Kim%
}{%
Doshi-Velez%
\ \BBA {} Kim%
}{%
{\protect \APACyear {2017}}%
}]{%
doshi2017towards}
\APACinsertmetastar {%
doshi2017towards}%
\begin{APACrefauthors}%
Doshi-Velez, F.%
\BCBT {}\ \BBA {} Kim, B.%
\end{APACrefauthors}%
\unskip\
\newblock
\APACrefYearMonthDay{2017}{}{}.
\newblock
{\BBOQ}\APACrefatitle {Towards a rigorous science of interpretable machine
  learning} {Towards a rigorous science of interpretable machine
  learning}.{\BBCQ}
\newblock
\APACjournalVolNumPages{arXiv preprint arXiv:1702.08608}{}{}{}.
\PrintBackRefs{\CurrentBib}

\bibitem [\protect \citeauthoryear {%
Esteva%
\ \protect \BOthers {.}}{%
Esteva%
\ \protect \BOthers {.}}{%
{\protect \APACyear {2017}}%
}]{%
esteva2017dermatologist}
\APACinsertmetastar {%
esteva2017dermatologist}%
\begin{APACrefauthors}%
Esteva, A.%
, Kuprel, B.%
, Novoa, R\BPBI A.%
, Ko, J.%
, Swetter, S\BPBI M.%
, Blau, H\BPBI M.%
\BCBL {}\ \BBA {} Thrun, S.%
\end{APACrefauthors}%
\unskip\
\newblock
\APACrefYearMonthDay{2017}{}{}.
\newblock
{\BBOQ}\APACrefatitle {Dermatologist-level classification of skin cancer with
  deep neural networks} {Dermatologist-level classification of skin cancer with
  deep neural networks}.{\BBCQ}
\newblock
\APACjournalVolNumPages{Nature}{542}{7639}{115--118}.
\PrintBackRefs{\CurrentBib}

\bibitem [\protect \citeauthoryear {%
Friedman%
\ \BBA {} Fisher%
}{%
Friedman%
\ \BBA {} Fisher%
}{%
{\protect \APACyear {1999}}%
}]{%
friedman1999bump}
\APACinsertmetastar {%
friedman1999bump}%
\begin{APACrefauthors}%
Friedman, J\BPBI H.%
\BCBT {}\ \BBA {} Fisher, N\BPBI I.%
\end{APACrefauthors}%
\unskip\
\newblock
\APACrefYearMonthDay{1999}{}{}.
\newblock
{\BBOQ}\APACrefatitle {Bump hunting in high-dimensional data} {Bump hunting in
  high-dimensional data}.{\BBCQ}
\newblock
\APACjournalVolNumPages{Statistics and Computing}{9}{2}{123--143}.
\PrintBackRefs{\CurrentBib}

\bibitem [\protect \citeauthoryear {%
Friedman%
\ \BBA {} Popescu%
}{%
Friedman%
\ \BBA {} Popescu%
}{%
{\protect \APACyear {2008}}%
}]{%
friedman2008predictive}
\APACinsertmetastar {%
friedman2008predictive}%
\begin{APACrefauthors}%
Friedman, J\BPBI H.%
\BCBT {}\ \BBA {} Popescu, B\BPBI E.%
\end{APACrefauthors}%
\unskip\
\newblock
\APACrefYearMonthDay{2008}{}{}.
\newblock
{\BBOQ}\APACrefatitle {Predictive learning via rule ensembles} {Predictive
  learning via rule ensembles}.{\BBCQ}
\newblock
\APACjournalVolNumPages{The Annals of Applied Statistics}{2}{3}{916--954}.
\PrintBackRefs{\CurrentBib}

\bibitem [\protect \citeauthoryear {%
Goh%
\ \BBA {} Rudin%
}{%
Goh%
\ \BBA {} Rudin%
}{%
{\protect \APACyear {2014}}%
}]{%
goh2014box}
\APACinsertmetastar {%
goh2014box}%
\begin{APACrefauthors}%
Goh, S\BPBI T.%
\BCBT {}\ \BBA {} Rudin, C.%
\end{APACrefauthors}%
\unskip\
\newblock
\APACrefYearMonthDay{2014}{}{}.
\newblock
{\BBOQ}\APACrefatitle {Box drawings for learning with imbalanced data} {Box
  drawings for learning with imbalanced data}.{\BBCQ}
\newblock
\BIn{} \APACrefbtitle {Proceedings of the 20th ACM SIGKDD international
  conference on Knowledge discovery and data mining} {Proceedings of the 20th
  acm sigkdd international conference on knowledge discovery and data mining}\
  (\BPGS\ 333--342).
\PrintBackRefs{\CurrentBib}

\bibitem [\protect \citeauthoryear {%
Goodman%
\ \BBA {} Flaxman%
}{%
Goodman%
\ \BBA {} Flaxman%
}{%
{\protect \APACyear {2017}}%
}]{%
goodman2017european}
\APACinsertmetastar {%
goodman2017european}%
\begin{APACrefauthors}%
Goodman, B.%
\BCBT {}\ \BBA {} Flaxman, S.%
\end{APACrefauthors}%
\unskip\
\newblock
\APACrefYearMonthDay{2017}{}{}.
\newblock
{\BBOQ}\APACrefatitle {European {U}nion regulations on algorithmic
  decision-making and a “right to explanation”} {European {U}nion
  regulations on algorithmic decision-making and a “right to
  explanation”}.{\BBCQ}
\newblock
\APACjournalVolNumPages{AI magazine}{38}{3}{50--57}.
\PrintBackRefs{\CurrentBib}

\bibitem [\protect \citeauthoryear {%
Guidotti%
\ \protect \BOthers {.}}{%
Guidotti%
\ \protect \BOthers {.}}{%
{\protect \APACyear {2018}}%
}]{%
guidotti2018survey}
\APACinsertmetastar {%
guidotti2018survey}%
\begin{APACrefauthors}%
Guidotti, R.%
, Monreale, A.%
, Ruggieri, S.%
, Turini, F.%
, Giannotti, F.%
\BCBL {}\ \BBA {} Pedreschi, D.%
\end{APACrefauthors}%
\unskip\
\newblock
\APACrefYearMonthDay{2018}{}{}.
\newblock
{\BBOQ}\APACrefatitle {A survey of methods for explaining black box models} {A
  survey of methods for explaining black box models}.{\BBCQ}
\newblock
\APACjournalVolNumPages{ACM computing surveys (CSUR)}{51}{5}{1--42}.
\PrintBackRefs{\CurrentBib}

\bibitem [\protect \citeauthoryear {%
Han%
, Pei%
\BCBL {}\ \BBA {} Yin%
}{%
Han%
\ \protect \BOthers {.}}{%
{\protect \APACyear {2000}}%
}]{%
han2000mining}
\APACinsertmetastar {%
han2000mining}%
\begin{APACrefauthors}%
Han, J.%
, Pei, J.%
\BCBL {}\ \BBA {} Yin, Y.%
\end{APACrefauthors}%
\unskip\
\newblock
\APACrefYearMonthDay{2000}{}{}.
\newblock
{\BBOQ}\APACrefatitle {Mining frequent patterns without candidate generation}
  {Mining frequent patterns without candidate generation}.{\BBCQ}
\newblock
\APACjournalVolNumPages{ACM sigmod record}{29}{2}{1--12}.
\PrintBackRefs{\CurrentBib}

\bibitem [\protect \citeauthoryear {%
H{\"u}llermeier%
\ \BBA {} Waegeman%
}{%
H{\"u}llermeier%
\ \BBA {} Waegeman%
}{%
{\protect \APACyear {2021}}%
}]{%
hullermeier2021aleatoric}
\APACinsertmetastar {%
hullermeier2021aleatoric}%
\begin{APACrefauthors}%
H{\"u}llermeier, E.%
\BCBT {}\ \BBA {} Waegeman, W.%
\end{APACrefauthors}%
\unskip\
\newblock
\APACrefYearMonthDay{2021}{}{}.
\newblock
{\BBOQ}\APACrefatitle {Aleatoric and epistemic uncertainty in machine learning:
  An introduction to concepts and methods} {Aleatoric and epistemic uncertainty
  in machine learning: An introduction to concepts and methods}.{\BBCQ}
\newblock
\APACjournalVolNumPages{Machine Learning}{110}{3}{457--506}.
\PrintBackRefs{\CurrentBib}

\bibitem [\protect \citeauthoryear {%
Kohavi%
}{%
Kohavi%
}{%
{\protect \APACyear {1996}}%
}]{%
kohavi1996scaling}
\APACinsertmetastar {%
kohavi1996scaling}%
\begin{APACrefauthors}%
Kohavi, R.%
\end{APACrefauthors}%
\unskip\
\newblock
\APACrefYearMonthDay{1996}{}{}.
\newblock
{\BBOQ}\APACrefatitle {Scaling up the accuracy of naive-bayes classifiers: A
  decision-tree hybrid.} {Scaling up the accuracy of naive-bayes classifiers: A
  decision-tree hybrid.}{\BBCQ}
\newblock
\BIn{} \APACrefbtitle {Kdd} {Kdd}\ (\BVOL~96, \BPGS\ 202--207).
\PrintBackRefs{\CurrentBib}

\bibitem [\protect \citeauthoryear {%
Libbrecht%
\ \BBA {} Noble%
}{%
Libbrecht%
\ \BBA {} Noble%
}{%
{\protect \APACyear {2015}}%
}]{%
libbrecht2015machine}
\APACinsertmetastar {%
libbrecht2015machine}%
\begin{APACrefauthors}%
Libbrecht, M\BPBI W.%
\BCBT {}\ \BBA {} Noble, W\BPBI S.%
\end{APACrefauthors}%
\unskip\
\newblock
\APACrefYearMonthDay{2015}{}{}.
\newblock
{\BBOQ}\APACrefatitle {Machine learning applications in genetics and genomics}
  {Machine learning applications in genetics and genomics}.{\BBCQ}
\newblock
\APACjournalVolNumPages{Nature Reviews Genetics}{16}{6}{321--332}.
\PrintBackRefs{\CurrentBib}

\bibitem [\protect \citeauthoryear {%
Molnar%
}{%
Molnar%
}{%
{\protect \APACyear {2020}}%
}]{%
molnar2020interpretable}
\APACinsertmetastar {%
molnar2020interpretable}%
\begin{APACrefauthors}%
Molnar, C.%
\end{APACrefauthors}%
\unskip\
\newblock
\APACrefYear{2020}.
\newblock
\APACrefbtitle {Interpretable Machine Learning} {Interpretable machine
  learning}.
\newblock
\APACaddressPublisher{}{Lulu. com}.
\PrintBackRefs{\CurrentBib}

\bibitem [\protect \citeauthoryear {%
Murdoch%
, Singh%
, Kumbier%
, Abbasi-Asl%
\BCBL {}\ \BBA {} Yu%
}{%
Murdoch%
\ \protect \BOthers {.}}{%
{\protect \APACyear {2019}}%
}]{%
murdoch2019interpretable}
\APACinsertmetastar {%
murdoch2019interpretable}%
\begin{APACrefauthors}%
Murdoch, W\BPBI J.%
, Singh, C.%
, Kumbier, K.%
, Abbasi-Asl, R.%
\BCBL {}\ \BBA {} Yu, B.%
\end{APACrefauthors}%
\unskip\
\newblock
\APACrefYearMonthDay{2019}{}{}.
\newblock
{\BBOQ}\APACrefatitle {Interpretable machine learning: definitions, methods,
  and applications} {Interpretable machine learning: definitions, methods, and
  applications}.{\BBCQ}
\newblock
\APACjournalVolNumPages{arXiv preprint arXiv:1901.04592}{}{}{}.
\PrintBackRefs{\CurrentBib}

\bibitem [\protect \citeauthoryear {%
Rudin%
}{%
Rudin%
}{%
{\protect \APACyear {2019}}%
}]{%
rudin2019stop}
\APACinsertmetastar {%
rudin2019stop}%
\begin{APACrefauthors}%
Rudin, C.%
\end{APACrefauthors}%
\unskip\
\newblock
\APACrefYearMonthDay{2019}{}{}.
\newblock
{\BBOQ}\APACrefatitle {Stop explaining black box machine learning models for
  high stakes decisions and use interpretable models instead} {Stop explaining
  black box machine learning models for high stakes decisions and use
  interpretable models instead}.{\BBCQ}
\newblock
\APACjournalVolNumPages{Nature Machine Intelligence}{1}{5}{206--215}.
\PrintBackRefs{\CurrentBib}

\bibitem [\protect \citeauthoryear {%
Shen%
\ \protect \BOthers {.}}{%
Shen%
\ \protect \BOthers {.}}{%
{\protect \APACyear {2015}}%
}]{%
shen2015machine}
\APACinsertmetastar {%
shen2015machine}%
\begin{APACrefauthors}%
Shen, D.%
, Wu, G.%
, Zhang, D.%
, Suzuki, K.%
, Wang, F.%
\BCBL {}\ \BBA {} Yan, P.%
\end{APACrefauthors}%
\unskip\
\newblock
\APACrefYearMonthDay{2015}{}{}.
\newblock
{\BBOQ}\APACrefatitle {Machine learning in medical imaging.} {Machine learning
  in medical imaging.}{\BBCQ}
\newblock
\APACjournalVolNumPages{Comp. Med. Imag. and Graph.}{41}{}{1--2}.
\PrintBackRefs{\CurrentBib}

\bibitem [\protect \citeauthoryear {%
Van~Laarhoven%
\ \BBA {} Aarts%
}{%
Van~Laarhoven%
\ \BBA {} Aarts%
}{%
{\protect \APACyear {1987}}%
}]{%
van1987simulated}
\APACinsertmetastar {%
van1987simulated}%
\begin{APACrefauthors}%
Van~Laarhoven, P\BPBI J.%
\BCBT {}\ \BBA {} Aarts, E\BPBI H.%
\end{APACrefauthors}%
\unskip\
\newblock
\APACrefYearMonthDay{1987}{}{}.
\newblock
{\BBOQ}\APACrefatitle {Simulated annealing} {Simulated annealing}.{\BBCQ}
\newblock
\BIn{} \APACrefbtitle {Simulated annealing: Theory and applications} {Simulated
  annealing: Theory and applications}\ (\BPGS\ 7--15).
\newblock
\APACaddressPublisher{}{Springer}.
\PrintBackRefs{\CurrentBib}

\bibitem [\protect \citeauthoryear {%
Varshney%
\ \BBA {} Alemzadeh%
}{%
Varshney%
\ \BBA {} Alemzadeh%
}{%
{\protect \APACyear {2017}}%
}]{%
varshney2017safety}
\APACinsertmetastar {%
varshney2017safety}%
\begin{APACrefauthors}%
Varshney, K\BPBI R.%
\BCBT {}\ \BBA {} Alemzadeh, H.%
\end{APACrefauthors}%
\unskip\
\newblock
\APACrefYearMonthDay{2017}{}{}.
\newblock
{\BBOQ}\APACrefatitle {On the safety of machine learning: Cyber-physical
  systems, decision sciences, and data products} {On the safety of machine
  learning: Cyber-physical systems, decision sciences, and data
  products}.{\BBCQ}
\newblock
\APACjournalVolNumPages{Big data}{5}{3}{246--255}.
\PrintBackRefs{\CurrentBib}

\bibitem [\protect \citeauthoryear {%
Wachter%
, Mittelstadt%
\BCBL {}\ \BBA {} Floridi%
}{%
Wachter%
\ \protect \BOthers {.}}{%
{\protect \APACyear {2017}}%
}]{%
wachter17}
\APACinsertmetastar {%
wachter17}%
\begin{APACrefauthors}%
Wachter, S.%
, Mittelstadt, B.%
\BCBL {}\ \BBA {} Floridi, L.%
\end{APACrefauthors}%
\unskip\
\newblock
\APACrefYearMonthDay{2017}{06}{}.
\newblock
{\BBOQ}\APACrefatitle {{Why a Right to Explanation of Automated Decision-Making
  Does Not Exist in the General Data Protection Regulation}} {{Why a Right to
  Explanation of Automated Decision-Making Does Not Exist in the General Data
  Protection Regulation}}.{\BBCQ}
\newblock
\APACjournalVolNumPages{International Data Privacy Law}{7}{2}{76-99}.
\PrintBackRefs{\CurrentBib}

\bibitem [\protect \citeauthoryear {%
Wang%
\ \protect \BOthers {.}}{%
Wang%
\ \protect \BOthers {.}}{%
{\protect \APACyear {2015}}%
}]{%
wang2015bayesian}
\APACinsertmetastar {%
wang2015bayesian}%
\begin{APACrefauthors}%
Wang, T.%
, Rudin, C.%
, Doshi, F.%
, Liu, Y.%
, Klampfl, E.%
\BCBL {}\ \BBA {} MacNeille, P.%
\end{APACrefauthors}%
\unskip\
\newblock
\APACrefYearMonthDay{2015}{}{}.
\newblock
\APACrefbtitle {Bayesian or’s of and’s for interpretable classification
  with application to context aware recommender systems} {Bayesian or’s of
  and’s for interpretable classification with application to context aware
  recommender systems}\ \APACbVolEdTR{}{\BTR{}}.
\newblock
\APACaddressInstitution{}{MIT, Tech. Rep., 2015, submitted}.
\PrintBackRefs{\CurrentBib}

\bibitem [\protect \citeauthoryear {%
Wang%
\ \protect \BOthers {.}}{%
Wang%
\ \protect \BOthers {.}}{%
{\protect \APACyear {2017}}%
}]{%
wang2017bayesian}
\APACinsertmetastar {%
wang2017bayesian}%
\begin{APACrefauthors}%
Wang, T.%
, Rudin, C.%
, Doshi-Velez, F.%
, Liu, Y.%
, Klampfl, E.%
\BCBL {}\ \BBA {} MacNeille, P.%
\end{APACrefauthors}%
\unskip\
\newblock
\APACrefYearMonthDay{2017}{}{}.
\newblock
{\BBOQ}\APACrefatitle {A bayesian framework for learning rule sets for
  interpretable classification} {A bayesian framework for learning rule sets
  for interpretable classification}.{\BBCQ}
\newblock
\APACjournalVolNumPages{The Journal of Machine Learning
  Research}{18}{1}{2357--2393}.
\PrintBackRefs{\CurrentBib}

\end{thebibliography}
\newpage

\section*{Supporting information}

\begin{table}[h]
\begingroup 
\captionof{table}{Taxonomy}\label{tb:taxonomy}
\endgroup
\begin{adjustbox}{width=\columnwidth}
\begin{tabular}{ccll}
\toprule
$\{y_{i} = 1\ |\ \bR^+(\bx_{i}) = 1, \bR^-(\bx_{i}) = 0 \}$ & $=$ & Consensus - True Positive   & $(\contp)$,\\
$\{y_{i} = 0\ |\ \bR^+(\bx_{i}) = 1, \bR^-(\bx_{i}) = 0\}$ & $=$ & Consensus  - False Positive  & $(\confp)$,\\
$\{y_{i} = 0\ |\ \bR^+(\bx_{i}) = 0, \bR^-(\bx_{i}) = 1 \}$ & $=$ & Consensus  - True Negative  & $(\contn)$,\\
$\{y_{i} = 1\ |\ \bR^+(\bx_{i}) = 0, \bR^-(\bx_{i}) = 1 \}$ & $=$ & Consensus  - False Negative  & $(\confn)$,\\
$\{y_{i} = 1\ |\ \bR^+(\bx_{i}) = 1, \bR^-(\bx_{i}) = 1 \}$ & $=$ & Active Ambiguous - Positive  & $(\aap)$,\\
$\{y_{i} = 0\ |\ \bR^+(\bx_{i}) = 1, \bR^-(\bx_{i}) = 1 \}$ & $=$ & Active Ambiguous - Negative  & $(\aan)$,\\
$\{y_{i} = 1\ |\ \bR^+(\bx_{i}) = 0, \bR^-(\bx_{i}) = 0 \}$ & $=$ & Passive Ambiguous - Positive   & $(\pap)$,\\
$\{y_{i} = 0\ |\ \bR^+(\bx_{i}) = 0, \bR^-(\bx_{i}) = 0 \}$ & $=$ & Passive Ambiguous - Negative  & $(\pan)$.\\
\bottomrule
\end{tabular}
\end{adjustbox}
\end{table}

\begin{algorithm}[h]
\caption{Simulated annealing for finding MAP}\label{tb:algorithm}
\begin{algorithmic}[1]
\Procedure{SOAR}{$...$}

\State $\bR^+(0),\bR^-(0) \gets$ initial rule sets
\State $\big(\bR^+_*,\bR^-_*\big)\gets\big(\bR^+(0),\bR^-(0)\big)$
\State$t\gets 1$
\While {$t < iter_{max}$}
\State $\bx_i \gets $ a randomly selected observation with incorrect prediction (based on $\big(\bR^+_*,\bR^-_*\big)$)
\If {$\big[y_i==1\big]$}
\If {$\big[\bR^+_*(\bx_i)==1$ AND $\bR^-_*(\bx_i)==1\big]$}
    $\bR_{temp}^{\pm}\gets COVERLESS\Big(\bR^-(t-1)\Big)$
\EndIf
\If {$\big[\bR^+_*(\bx_i)==0$ AND $\bR^-_*(\bx_i)==0\big]$}
     $\bR_{temp}^{\pm}\gets COVERMORE\Big(\bR^+(t-1)\Big)$
\EndIf
\If {$\big[\bR^+_*(\bx_i)==0$ AND $\bR^-_*(\bx_i)==1\big]$}
    \[
        \bR_{temp}^{\pm}\gets\begin{cases}
        COVERLESS\Big(\bR^-(t-1)\Big)&\mbox{ with prob }0.5\\
        COVERMORE\Big(\bR^+(t-1)\Big)&\mbox{ with prob }0.5\\
        \end{cases}
    \]
\EndIf
\EndIf
\If {$\big[y_i==0\big]$}
\If {$\big[\bR^+_*(\bx_i)==1$ AND $\bR^-_*(\bx_i)==1\big]$}
    $\bR_{temp}^{\pm}\gets COVERLESS\Big(\bR^+(t-1)\Big)$
\EndIf
\If {$\big[\bR^+_*(\bx_i)==0$ AND $\bR^-_*(\bx_i)==0\big]$}
     $\bR_{temp}^{\pm}\gets COVERMORE\Big(\bR^-(t-1)\Big)$
\EndIf
\If {$\big[\bR^+_*(\bx_i)==1$ AND $\bR^-_*(\bx_i)==0\big]$}
    \[
        \bR_{temp}^{\pm}\gets\begin{cases}
        COVERLESS\Big(\bR^+(t-1)\Big)&\mbox{ with prob }0.5\\
        COVERMORE\Big(\bR^-(t-1)\Big)&\mbox{ with prob }0.5\\
        \end{cases}
    \]
\EndIf
\EndIf
\If {$\big[\op{Score}\big(\bR_{temp}^{\pm},\bR^{\mp}(t-1)\big)<\op{Score}(\bR^+_*,\bR^-_*)\big]$} $(\bR^+_*,\bR^-_*)\gets $
$\big(\bR_{temp}^{\pm},\bR^{\mp}(t-1)\big)$.
\EndIf
\State $\alpha\gets\min\Big\{1, \exp\Big(-\frac{\op{Score}\big(\bR_{temp}^{\pm},\bR^{\mp}(t-1)\big)-\op{Score}\,\big(\bR^+(t-1),\bR^-(t-1)\big)}{T(t)}\Big)\Big\}$
\State \[
        \bR^{\pm}(t)\gets\begin{cases}
        \bR_{temp}^{\pm}&\mbox{ with prob }\alpha\\
        \bR^{\pm}(t-1)&\mbox{ with prob }1-\alpha\\
        \end{cases}
    \]
\State $t\gets t+1$
\EndWhile
\State \textbf{return} $(\bR^+_*,\bR^-_*)$
\EndProcedure
\end{algorithmic}
\end{algorithm}

\begin{table}[h]
\begin{center}
\begingroup
\captionof{table}{Assignment of true classes for creation of synthetic data.}\label{tb:synth}
\endgroup
\scalebox{0.9}{
\begin{tabular}{ccc}
\toprule
    {\bf Initial class} & {\bf Count} & {\bf Final class}\\
    \midrule
    positive only  & 441 & positive\\
    negative only &  431 & negative\\
    both positive and negative & 59 & 5\% positive, 95\% negative\\
    neither & 69 & 50\% positive, 50\% negative\\
    \bottomrule
\end{tabular}
}
\end{center}
\end{table}

\begin{table}[h]
\begin{center}
\begingroup
\captionof{table}{True class and predicted class for all $2^5$ possible predictor combinations.}\label{tb:synth.fin}
\endgroup
\scalebox{0.8}{
\begin{tabular}{ccccccc}
\toprule
    {\bf x1} & {\bf x2} & {\bf x3} & {\bf x4} & {\bf x5} & {\bf Truth} & {\bf Prediction} \\
    \midrule
    0 & 0 & 0 & 0 & 0 & (0,1) & (0,1)\\
    0 & 0 & 0 & 0 & 1 & (1,0) & (1,0)\\
    0 & 0 & 0 & 1 & 0 & (0,0) & (0,1)*\\
    0 & 0 & 0 & 1 & 1 & (1,0) & (1,0)\\
    0 & 0 & 1 & 0 & 0 & (0,1) & (0,1)\\
    0 & 0 & 1 & 0 & 1 & (0,1) & (0,0)*\\
    0 & 0 & 1 & 1 & 0 & (0,0) & (0,1)*\\
    0 & 0 & 1 & 1 & 1 & (0,0) & (0,0)\\
    0 & 1 & 0 & 0 & 0 & (1,0) & (1,0)\\
    0 & 1 & 0 & 0 & 1 & (1,0) & (1,0)\\
    0 & 1 & 0 & 1 & 0 & (1,0) & (1,0)\\
    0 & 1 & 0 & 1 & 1 & (1,0) & (1,0)\\
    0 & 1 & 1 & 0 & 0 & (1,0) & (1,0)\\
    0 & 1 & 1 & 0 & 1 & (1,0) & (1,0)\\
    0 & 1 & 1 & 1 & 0 & (1,0) & (1,0)\\
    0 & 1 & 1 & 1 & 1 & (1,0) & (1,0)\\
    
    1 & 0 & 0 & 0 & 0 & (0,1) & (0,1)\\
    1 & 0 & 0 & 0 & 1 & (1,1) & (1,1)\\
    1 & 0 & 0 & 1 & 0 & (0,1) & (0,1)\\
    1 & 0 & 0 & 1 & 1 & (1,1) & (1,1)\\
    1 & 0 & 1 & 0 & 0 & (0,1) & (0,1)\\
    1 & 0 & 1 & 0 & 1 & (0,1) & (0,1)\\
    1 & 0 & 1 & 1 & 0 & (0,1) & (0,1)\\
    1 & 0 & 1 & 1 & 1 & (0,1) & (0,1)\\
    1 & 1 & 0 & 0 & 0 & (1,0) & (1,0)\\
    1 & 1 & 0 & 0 & 1 & (1,0) & (1,0)\\
    1 & 1 & 0 & 1 & 0 & (1,0) & (1,0)\\
    1 & 1 & 0 & 1 & 1 & (1,0) & (1,0)\\
    1 & 1 & 1 & 0 & 0 & (0,1) & (1,1)*\\
    1 & 1 & 1 & 0 & 1 & (0,1) & (1,1)*\\
    1 & 1 & 1 & 1 & 0 & (0,1) & (1,1)*\\
    1 & 1 & 1 & 1 & 1 & (0,1) & (1,1)*\\
    \bottomrule
\end{tabular}}
\end{center}
\end{table}

\begin{table}[h]
\begin{center}
\begingroup
\captionof{table}{Comparison of performance of SOAR, BOA and random forest on synthetic data.}\label{tb:synth.res}
\endgroup
\scalebox{0.8}{
\begin{tabular}{ccccc}
\toprule
     & {\bf SOAR} & {\bf SOAR} (forced) & {\bf BOA} & {\bf RF}\\
     \midrule
    \makecell{truly misclassified} & 0 & 0.005 & 0.05 & 0.05\\ 
    \makecell{ambiguous} & 0.25 & 0.07 & 0 & 0 \\
    \bottomrule
\end{tabular}
}
\end{center}
\end{table}

\begin{table}[h]
\begin{center}
\begingroup
\captionof{table}{Comparison of performances of SOAR, BOA and random forest using Adult data.}\label{tb:adult.comp}
\endgroup
\scalebox{0.8}{
\begin{tabular}{ccccc}
\toprule
     & {\bf SOAR} & {\bf SOAR} (forced) & {\bf BOA} & {\bf RF}\\
     \midrule
    \makecell{truly misclassified} & 0.034 & 0.174 & 0.297 & 0.159\\ 
    \makecell{ambiguous} & 0.432 & 0.044 & 0 & 0 \\ 
    \bottomrule

\end{tabular}
}
\end{center}
\end{table}

\begin{table}[h]
\begin{center}
\begingroup
\captionof{table}{Examples of cases from Adult data labeled as {\it ambiguous} by SOAR algorithm.}\label{tb:adult.ex}
\endgroup
\scalebox{0.55}{
\begin{tabular}{ccccccccccc}
\toprule
    & \rotatebox{90}{{\bf education}} & \rotatebox{90}{{\bf ed\_num}} & \rotatebox{90}{{\bf marital\_status}} & \rotatebox{90}{{\bf occupation}} & \rotatebox{90}{{\bf relationship}} & \rotatebox{90}{{\bf race}} & \rotatebox{90}{{\bf sex}} & \rotatebox{90}{{\bf capital\_loss}} & \rotatebox{90}{{\bf capital\_gain}} & \rotatebox{90}{{\bf $>$50K}}\\
    \midrule
    \multirow{3}{*}{\rotatebox{90}{ $\substack{\mbox{\large Active} \\ \mbox{\large ambiguous}}$ }} \Gape[10pt] &  Masters  &  14 & Married-civ-spouse & Other-service & Husband & White & Male &  0        &  0  &   0\\
    \Gape[10pt] & Bachelors & 13 & Never-married & Exec-managerial & Own-child & White  & Male  & 0 & 0  & 0\\
    \Gape[10pt] & Masters & 14 & Never-married   & Sales & Own-child & Black & Male & 0  & 0 & 0\\
    \Gape[4pt] &  &  &   & &  &  &  &  &  & \\
    \multirow{3}{*}{\rotatebox{90}{ $\substack{\mbox{\large Passive} \\ \mbox{\large ambiguous}}$ }} \Gape[10pt] &  Bachelors            & 13 & Never-married & Exec-managerial & Other-relative & White & Female & 0  & 1  & 0\\
    \Gape[10pt] & 10th & 6 & Never-married & Adm-clerical   & Not-in-family & White & Female & 1 & 0 & 0\\
    \Gape[10pt] & Masters & 14 & Divorced  & Exec-managerial & Not-in-family & White & Male & 0 & 1 & 1\\
    \bottomrule
\end{tabular}}
\end{center}
\end{table}

\begin{table}[h]
\begin{center}
\begingroup
\captionof{table}{Comparison between traditional prediction classification for BOA and the newly proposed SOAR taxonomy using Adult data.}\label{tb:adult.tax}
\endgroup
\scalebox{0.9}{
\begin{tabular}{ccccccccc}
\toprule
     & $\contp$ & $\contn$ & $\confn$ & $\confp$ & $\aan$ & $\aap$ & $\pan$ & $\pap$\\
     \midrule
     \makecell{$\tp$} & 433 & 0 & 21 & 0 & 0 & 1234 & 0 & 39\\
     \makecell{$\tn$} & 0 & 3525  &  0   & 0  & 16 &   0  & 45   & 0  \\
     \makecell{$\fn$} & 0  &  0 & 137   & 0  &  0  &  1  &  0 &  11\\
     \makecell{$\fp$} & 0  & 68 &   0 & 118 & 1864  &  0  & 49  &  0\\
     \bottomrule
\end{tabular}
}
\end{center}
\end{table}

\begin{table}[h]
\begin{center}
\begingroup
\captionof{table}{Most common rules for Adult data.}\label{tb:adult.freq}
\endgroup
    \scalebox{0.9}{
    \begin{tabular}{ccc}
    \toprule
        {\bf Classification} & {\bf Rule} & {\bf Frequency}\\
        \midrule
        \multirow{3}{*}{Positive}
         & (marital\_status, Married-civ-spouse) & 42\\
         & (education\_num, 15) & 38\\
         & (education, Prof-school) & 37\\
         & & \\
         \multirow{3}{*}{Negative} & (relationship, Own-child) & 54 \\
          & (occupation, Other-service)  & 42\\
          & (capital\_gain, 0) $\wedge$ (capital\_loss, 0)  & 41\\
          \bottomrule
    \end{tabular}}
\end{center}
\end{table}

\begin{table}[htbp]
\begin{center}
\begingroup
\captionof{table}{Comparison of performances of SOAR, BOA and random forest using car data.}\label{tb:car.comp}
\endgroup
\scalebox{0.8}{
\begin{tabular}{ccccc}
\toprule
     & {\bf SOAR} & {\bf SOAR} (forced) & {\bf BOA} & {\bf RF}\\
     \midrule
    \makecell{truly misclassified} & 0.006 & 0.009 & 0.153 & 0.013\\ 
    \makecell{ambiguous} & 0.242 & 0.235 & 0 & 0 \\
    \bottomrule
\end{tabular}
}
\end{center}
\end{table}

\begin{table}[h]
\begin{center}
\begingroup
\captionof{table}{Examples of cases from car data labeled as {\it ambiguous} by SOAR algorithm.}\label{tb:car.ex}
\endgroup
\scalebox{0.7}{
    \begin{tabular}{cccccccc}
    \toprule
    \Gape[10pt]{SOAR} & {\bf buyingPrice} &   {\bf maintainPrice} & {\bf doors} &   {\bf persons} & {\bf lugBoot} & {\bf safety} & {\bf class}   \\
    \midrule
    \multirow{5}{*}{\rotatebox{90}{Active ambiguous}} 
    \Gape[10pt] & vhigh & low & 2 & 4 & small  & high & 1  \\
    \Gape[10pt] & vhigh & low & 2 & more & small  & high & 0  \\
    \Gape[10pt] & high & low & 2 & more & small  & high & 0  \\
    \Gape[10pt] & low & low & 2 & 4 & small  & high & 1  \\
    \Gape[5pt] &  &  &  &  &  &  & \\
    \multirow{5}{*}{\rotatebox{90}{Passive ambiguous}} \Gape[10pt] & vhigh & med & 2 & 4 & big  & med & 1  \\
    \Gape[10pt] & low & low & 3 & more & med  & med & 1  \\
    \Gape[10pt] & low & med & 5more & more & big  & high & 1  \\
    \Gape[10pt] & high & low & 4 & more & small &  med & 0\\
    \bottomrule
    \end{tabular}
}
\end{center}
\end{table}

\end{document}